\documentclass[journal]{IEEEtran}
\usepackage{amssymb}
\usepackage{pifont}
\usepackage{graphicx} 
\usepackage{amsmath} 
\usepackage{float}
\usepackage{amsfonts} 
\usepackage{pgfplots} 
\pgfplotsset{compat=1.18}
\usetikzlibrary{patterns}
\usepackage{multirow}
\usepackage{booktabs}
\usepackage{multirow}
\usepackage{array}
\usepackage{cite} 
\usepackage{booktabs}   
\usepackage{makecell}   
\usepackage{xcolor}

\ifCLASSINFOpdf
\else
\fi
\begin{document}
%
\title{MATS: A novel multi-modality multi-task learning framework for 3D perception in autonomous driving}
%
%

\author{Junchen Huo, Wanming Hao~\IEEEmembership{Senior Member,~IEEE}, Song Wang*, Enqing Chen, Shouyi Yang, and Guanghui Wang~\IEEEmembership{Senior Member,~IEEE}

\thanks{Junchen Huo, Wanming Hao, Song Wang and Enqing Chen are with the School of Electrical and Information Engineering, Zhengzhou University, Zhengzhou 450001, China. Shouyi Yang is with Tianping College of Suzhou University of Science and Technology, Suzhou 215009, China. Guanghui Wang is with the department of computer science, Toronto Metropolitan University, Toronto M5B2K3, Canada.}
\thanks{*Corresponding author (Song Wang, E-mail: wangsong61@163.com).}
}

\markboth{Journal of \LaTeX\ Class Files,~Vol.~14, No.~8, August~2015}%
{Shell \MakeLowercase{\textit{et al.}}: Bare Demo of IEEEtran.cls for IEEE Journals}

\maketitle

\begin{abstract}
Multi-modality data from different sensors provides rich complementary information for 3D perception, becoming an essential component in reliable autonomous driving systems. Current research typically designs intricate and complex fusion strategies to integrate information from multimodal data on a unified bird’s-eye-view (BEV) feature map for the joint learning of multiple perception tasks. However, such a single feature map hardly carries sufficient information to simultaneously meet the requirements of various perception tasks, leading to a very limited perception performance. To mitigate this limitation, this paper proposes MATS, a novel multi-modality multi-task learning approach with modality-adaptive BEV fusion and task-specific Mixture-of-Experts (MoE) for 3D perception. Specifically, a simple modality-adaptive BEV fusion module is designed to adaptively recalibrate the BEV features by modeling the global cross-modality dependencies, generating diverse BEV feature maps for various perception tasks. For joint multi-task learning, this paper proposes a task-specific MoE module to decouple the tasks and enable the network to automatically choose the appropriate BEV feature candidates for each specific task. To validate the effectiveness of the proposed approach, we conduct extensive experiments on the large-scale benchmark nuScenes. With the camera- and LiDAR-modality input data, the proposed approach outperforms the state-of-the-art (SOTA) by a significant margin. Furthermore, the experimental results on the single tasks show that the proposed approach significantly outperforms the baselines. The code and trained models will be available upon publication.

\end{abstract}


%
\IEEEpeerreviewmaketitle

\section{Introduction}
\label{intro}
%
%
%
%
\IEEEPARstart{3}{D} environment perception around self-driving vehicles is indispensable for autonomous driving systems\cite{alemaw2025modeling,wang2026privacy}. To accurately perceive the surrounding environment, current self-driving vehicles are typically equipped with advanced sensors, such as cameras, LiDAR and Radar. Among these, cameras have become the most popular choice due to their low cost and rapid advances in vision-based techniques. However, camera-based perception is highly sensitive to lighting conditions, including low illumination at night \cite{shi2024nitedr}, the reflections and shadows on the road \cite{wang2025ubtransformer}, or even the attack light signals from criminals \cite{guesmi2023physical}. To mitigate these risks, multi-modality perception that combines cameras with LiDAR has been proposed as a robust solution for 3D environment understanding \cite{zhao2023lif,zhang2025synet}. In this approach, cameras provide rich visual cues, while LiDAR delivers accurate 3D positional information, making their fusion complementary. This multi-modality solution has been widely adopted in autonomous driving systems by leading manufacturers such as Toyota and Volkswagen \cite{xiang2023multi}.

Due to the view discrepancy between cameras and LiDAR sensors mounted on the same vehicle, aligning their raw data at the pixel or point level is impractical. To address this challenge, recent methods commonly project multimodal data into a unified bird’s-eye-view (BEV) representation space \cite{huo2026cgmae}, where camera–LiDAR features can be aligned and integrated. Through an image-guided query initialization strategy, TransFusion\cite{bai2022transfusion} integrates 2D image features into LiDAR BEV features to locate the hard 3D objects in point clouds.
MapFusion \cite{hao2025mapfusion} leverages complementary information from BEV features in different modalities to enhance the feature representation for the map construction task. However, current task-specific multimodal learning approaches require autonomous driving systems to deploy multiple independent full-scene perception models, significantly increasing both computational cost and deployment complexity.


Multi-task learning offers a promising solution to achieve several perception tasks simultaneously. BEVFusion \cite{liu2023bevfusion} unifies multi-modality fusion and multi-task learning within an end-to-end framework, directly applying multi-task heads to acquire the location and semantic features of the objects from the shared camera-LiDAR BEV representation. MTA \cite{ma2024mta} is proposed to enforce the alignment between modalities through multimodal contextual learning and cross-modal prompting mechanisms, aiming to enhance the performance of 3D detection and 3D dense captioning. Chen et al. introduce M3Net \cite{chen2025m3net} to jointly predict 3D detection, segmentation, and occupancy results by integrating multi-modality features. Despite lower training and deployment costs, these end-to-end multi-modality multi-task learning methods often suffer from negative transfer, where the performance of joint-learning tasks is inferior to that of separately trained models \cite{liu2023bevfusion,huang2023fuller,xie2022m}.
This limitation significantly restricts the practical adoption of multi-task models in autonomous driving systems.

To alleviate the negative transfer problem, several research works have been proposed for multi-modality multi-task 3D perception. Following the baseline BEVFusion \cite{liu2023bevfusion}, FULLER\cite{huang2023fuller} is proposed to calibrate the gradients of different task losses, balancing the contributions of different tasks to the model optimization. From the perspective of feature-fusion, a multi-modal Transformer encoder is designed in UniTR \cite{wang2023unitr} to bridge multi-sensor data and map tokens to the BEV space. Zhao et al. introduce a unified Transformer decoder, i.e., MaskBEV \cite{zhao2024maskbev}, to disassemble the useful information of the BEV map for 3D object detection and map segmentation tasks. Despite these efforts, most of the existing methods rely on intricate fusion strategies that compress complementary multimodal information into a single BEV feature map. This design often fails to provide sufficient representational capacity for diverse perception tasks, resulting in only marginal improvements against negative transfer. In contrast, this paper argues that a simpler fusion strategy combined with diverse BEV feature maps can more effectively enhance 3D perception performance. Importantly, maintaining multiple BEV feature maps imposes only minimal memory overhead due to their inherently low resolution.

Therefore, this paper proposes a novel multi-modality multi-task learning approach (MATS) with modality-adaptive BEV fusion and task-specific Mixture-of-Experts (MoE) for 3D object detection and map segmentation. Specifically, the modality-adaptive BEV fusion module recalibrates the multi-modality BEV features by modeling the global cross-modal dependencies and generates multiple fused BEV feature maps for the fusion diversity. To support the joint multi-task learning, this paper designs a task-specific MoE module to decouple the multiple tasks and enable the network to automatically choose the appropriate fused BEV feature maps for each specific task. To validate the effectiveness of the proposed approach, we conduct extensive experiments on the large-scale benchmark nuScenes \cite{caesar2020nuscenes}. With the camera- and LiDAR-modality input data, the proposed approach outperforms the state-of-the-art (SOTA) by a significant margin. Furthermore, the experimental results on single tasks show that the proposed approach achieves better performance compared to the baselines. The contributions of this paper are summarized below.

\begin{enumerate}
\item This paper proposes a novel multi-modality multi-task learning approach with modality-adaptive BEV fusion and task-specific Mixture-of-Experts (MoE) for 3D object detection and map segmentation.

\item A modality-adaptive BEV fusion module is introduced to enhance the diversity of the BEV feature maps by recalibrating the multi-modality BEV features adaptively and generating multiple fused BEV candidates.

\item For joint multi-task learning, a task-specific MoE module is designed to decouple the multiple tasks and enable the network to automatically choose the appropriate fused BEV feature maps for each task.

\item This paper validates the effectiveness of the proposed approach on the large-scale nuScenes benchmark. Experimental results demonstrate that our method outperforms SOTA approaches by a significant margin. In addition, when evaluated on individual tasks, the proposed approach also achieves superior performance compared to existing baselines.

\end{enumerate}

The remainder of this paper is organized as follows. Section \ref{relatedworks} reviews related work on multi-modality fusion and multi-task learning for 3D perception. Section \ref{method} introduces the proposed multi-modality multi-task learning approach. Section \ref{experiments} reports the experimental results on the nuScenes benchmark, and Section \ref{conclusion} concludes the paper.

\section{Related Work}
\label{relatedworks}

In this section, we introduce the multi-modality fusion and multi-task learning methods for 3D perception.

\subsection{Multi-modality Fusion}

In recent years, multi-modality sensor fusion has emerged as a core approach to improve the performance of 3D environment perception in autonomous driving. Heterogeneous sensors such as camera, LiDAR, and Radar \cite{guo2020deep,patole2017automotive} provide distinct advantages. Cameras deliver rich semantic details, while LiDAR offers accurate range measurements and geometric structure. In the meantime, Radar exhibits the highest robustness under adverse weather and illumination conditions.



 To leverage these advantages, several fusion algorithms \cite{agrawal2023static, wang2020high, li2022mathsf, ravindran2022camera} integrate data from all three sensors to enhance perceptual capabilities. For example, Agrawal et al. \cite{agrawal2023static} propose a self-calibration method to achieve alignment and fusion of the data from three sensors. Wang et al. \cite{wang2020high} encode the acquired camera, LiDAR and Radar data and perform point-wise fusion of the data. CLR-BNN \cite{ravindran2022camera} is designed to apply a Bayesian neural network to realize the joint fusion of the three modalities. While these fusion algorithms enhance perception performance by integrating all three modalities, they often require substantial computational resources and result in longer response times.

\begin{figure*}[t]
    \centering
    \includegraphics[width=1\linewidth]{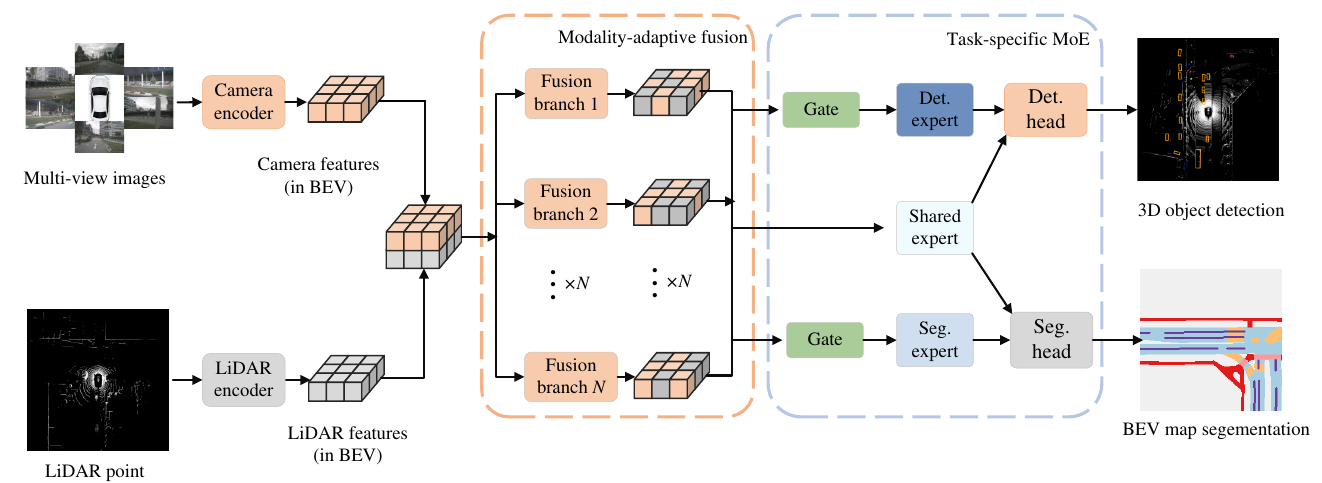}
    \caption{Illustration of the overall architecture of the proposed MATS. The multi-modality inputs from camera and LiDAR sensors are processed through the corresponding encoders. The modality-adaptive fusion module generates $N$ fused BEV feature maps with diverse representations. To leverage these representations, the Task MoE module splits the experts of MoE into the specialized experts and the shared ones. The separate task heads are utilized to predict the results of 3D object detection and BEV map segmentation.}
    \label{fig:framework}
\end{figure*}

 To mitigate the challenge, existing studies generally adopt two-modality data to fuse. MT-DETR \cite{Chu_2023_WACV} leverages a spatiotemporal scene network for adaptive camera-Radar data fusion. RCM-Fusion \cite{kim2024rcm} achieves camera-Radar fusion at two levels, i.e., the feature level and the instance level, where the multi-modal data are mapped into the BEV representation space to perform the feature-level fusion, while the instance-level fusion leverages radar grid point refinement. RCBEV \cite{zhou2023bridging} combines spatiotemporal encoding of millimetre-wave radar with multi-scale image features and adopts a two-stage fusion framework to achieve high-precision detection in challenging environments. SimpleBEV \cite{harley2022simple} achieves high-precision camera-Radar fusion in the BEV space by correcting camera depth estimation with a cascade network and optimizing radar features via multi-scale sparse convolution. SA-STNet \cite{hazarika2025multifusionnet} introduces a multi-scale attention mechanism to fuse the camera-Radar data. In RCBEVDet \cite{lin2024rcbevdet}, a cross-attention multi-layer fusion module is designed to automatically align the BEV features of Radar and camera. These camera-Radar fusion approaches yield a great performance improvement in 3D perception, but the low-resolution point cloud data captured by Radar sensors cannot perform precise mapping. Therefore, current approaches favor the integration of camera and LiDAR sensors \cite{bai2022transfusion,hao2025mapfusion,liu2023bevfusion,ma2024mta,fu2024eliminating,wang2023unitr}. ECfusion \cite{fu2024eliminating} aligns sensor information in the BEV space through a flow-based semantic alignment and query recovery mechanism. BEVFusion \cite{liu2023bevfusion} efficiently unifies geometric and semantic information within the BEV space by leveraging attention mechanisms. Building on cross-attention, TransFusion \cite{bai2022transfusion} adaptively fuses sparse object queries from LiDAR with image features to mitigate the degradation in image quality. MapFusion \cite{hao2025mapfusion} adopts dual-dynamic fusion to fully exploit useful information across modalities. Moreover, UnitR\cite{wang2023unitr} jointly models data from different modalities through parameter sharing on a unified multi-modal backbone.

 Although the above methods have demonstrated strong environmental perception capabilities, most are limited to single-task learning, requiring multiple independent models to achieve a comprehensive perception of the surrounding environment. In contrast, this paper proposes a multi-modality multi-task learning framework that performs 3D object detection and map segmentation simultaneously, thereby reducing computational cost and simplifying deployment in real-world applications.

\subsection{Multi-task learning}
Multi-task learning trains a model to address multiple related tasks simultaneously using shared data and representations. In the context of 3D perception, multi-task models can capture more comprehensive information about the surrounding environment than their single-task counterparts, which has sparked significant interest in the autonomous driving community \cite{liu2023bevfusion,huang2023fuller,chen2025m3net,zhao2024maskbev,xia2024henet,xiao2025adversarial}. M$^{2}$BEV \cite{xie2022m} projects multi-view image features into a shared BEV representation to jointly perform 3D object detection and map segmentation.  BEVFusion \cite{liu2023bevfusion} is designed to fuse camera-LiDAR data for joint learning of 3D object detection and map segmentation. Furthermore, M3Net \cite{chen2025m3net} extends multi-task learning to the occupancy task and proposes a learning framework for the 3D detection, segmentation, and occupancy tasks. These methods enable joint training of multiple tasks, but they often suffer from task conflicts, where the performance of the joint-training models is inferior to that of the single-training ones. Recent studies have attempted to address the negative transfer issue from different perspectives. HENet \cite{xia2024henet} utilizes independent BEV encoders and task decoders to decouple the joint task. FULLER \cite{huang2023fuller} calibrates task gradients at the final layer of the backbone network to alleviate task conflicts. MaskBEV\cite{zhao2024maskbev} designs a unified transformer decoder to disassemble the useful information from the BEV map for 3D object detection and map segmentation tasks. 

Nevertheless, existing multi-task learning methods for 3D perception often rely on a single BEV feature map to decode task-relevant information. However, such a representation rarely meets the diverse requirements of multiple tasks. In joint learning of 3D object detection and map segmentation, the 3D object detection task primarily focuses on accurately localizing individual objects where instance-level geometric features are required \cite{li2024quadbev,xia2024henet}. In contrast, the map segmentation task aims to capture the continuous semantic layout of the global scene \cite{huang2023fuller,hao2025msc}. Such a discrepancy in task requirements leads to severe feature conflicts within a single shared BEV representation. In this paper, diverse BEV feature maps are generated to alleviate the conflicts.



\section{Method}
\label{method}

This section presents the details of the proposed Multi-modality Multi-task Learning framework (MATS) for 3D object detection and map segmentation. Section \ref{overall} introduces the overall architecture of the framework. Section \ref{fusion} and Section \ref{moe} then provide detailed descriptions of the modality-adaptive BEV fusion module and the task-specific Mixture-of-Experts (MoE) module, respectively. Finally, Section \ref{loss} outlines the design of the task heads and the corresponding loss functions.

\subsection{Overall architecture}
\label{overall}
The overall architecture of MATS is illustrated in Fig. \ref{fig:framework}. Given multi-modality inputs, their corresponding encoders extract the features of the surrounding environment and map them into the BEV space. To enrich the carriers of the complementary information from multi-modality data, the modality-adaptive BEV fusion module is utilized to generate various fused BEV feature maps. After that, the task-specific MoE module splits the experts of MoE into the specialized experts to learn the focused knowledge and the shared ones to capture the common knowledge across different tasks. Finally, separate task heads are used to predict the detection and segmentation results simultaneously.

\begin{figure}[t]
    \centering
    \includegraphics[width=1\linewidth]{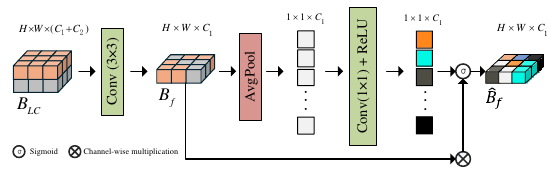}
    \caption{Illustration of the simple BEV fusion network utilized in the modality-adaptive module.}
    \label{fig:dynamic}
\end{figure}

\subsection{Modality-adaptive BEV fusion} 
\label{fusion}
In current multi-modality multi-task learning methods, a unified fused BEV feature map is typically shared by all the tasks. However, such a feature map hardly meets the requirements of different tasks for the various task-specific information. For example, the 3D object detection task focuses on the locations and the classes of the dynamic objects, whereas map segmentation tends to understand the scene layout, like the drivable and non-drivable areas. To enrich the information of the unified BEV feature map, existing methods are likely to design intricate and complex fusion strategies to integrate the complementary information from multi-modality data \cite{huang2023fuller,ma2024mta,yu2025sgformer}. Unlike previous methods, we employ a lightweight fusion network that generates multiple fused BEV feature maps, serving as diverse information carriers for different tasks.
Without assuming prior knowledge of the downstream tasks, the proposed modality-adaptive BEV fusion module is composed of multiple independent fusion branches. It is worth noting that the multi-branch BEV fusion in this paper refers to multiple independent fusion branches at the fusion stage, which is different from the use of multiple encoders for the inputs from different modalities in the existing methods \cite{song2024graphbev,kuang2024cmgfa}. In this paper, the branches are equipped with the same architecture and separate weights, thereby enabling the parallel generation of diverse fused BEV feature maps.
 The detailed architecture of the fusion branch is presented in Fig. \ref{fig:dynamic}.

As shown in Fig. \ref{fig:dynamic}, the fusion branch takes the concatenated camera BEV feature map $B_{Camera} \in \mathbb{R}^{H \times W \times C_1 }$ and the LiDAR BEV one $B_{Lidar} \in \mathbb{R}^{H \times W \times C_2 }$ as the input $B_{LC} \in \mathbb{R}^{H \times W \times (C_1+C_2) }$. We employ a simple $3\times3$ convolution layer to fuse the BEV feature maps between the camera and LiDAR, generating a coarse fusion result $B_f\in \mathbb{R}^{H \times W \times C_1 }$. Here,we equate the channel dimensions between the camera-based map and the fused map to utilize mature image processing solutions for further feature integration.
The coarse fusion map $B_f$ passes through a global average pooling layer \cite{hsiao2019filter} to aggregate the feature map across spatial dimensions, generating channel-wise feature responses $S$ with global spatial receptiveness on $B_f$. The following $1\times1$ convolution layer \cite{pang2017convolution} and ReLU layer \cite{agarap2018deep} are designed to learn a global nonlinear interaction between channels of $S$ and produce the recalibration vector $\hat{S}$. Finally, the fine fusion map $\hat{B_f}$ is obtained by reweighting $B_f$ using the channel-wise multiplication with the recalibration vector $\hat{S}$. The whole process of the fusion branch can be formulated as
\begin{equation} 
\label{eq:fusion}
\hat{B_f} = (Conv_{1\times1}(AvgPool(B_f)))\cdot B_f,
\end{equation}  
where $B_f = Conv_{3\times3}(B_{LC})$. With the multiple independent integration branches, the modality-adaptive BEV fusion module adaptively selects valuable information from two modalities to fuse and generates multiple diverse fused BEV feature maps for downstream networks.

\subsection{Task-specific MoE module}
\label{moe}
In joint multi-task learning, the inherent conflicts between different tasks are likely to harm the 3D perception performance, leading to worse results than the isolated task learning \cite{liu2023bevfusion,huang2023fuller,xie2022m}. To address this problem, the proposed task-specific MoE module adopts task-conditioned routing to assign features to the corresponding expert groups according to their task types, thereby avoiding task conflicts that would otherwise arise when a single shared gate mixes inconsistent optimization targets. Furthermore, the experts are partitioned into detection and segmentation groups, where each expert updates only its own parameters. This partitioning design isolates multi-task gradients within separate branches and mitigates multi-task gradient conflicts in the decoder.


Given the input $x$, the original MoE \cite{jacobs1991adaptive} can be formulated as Eq. (\ref{eq:moe_o}). 
\begin{equation}
\label{eq:moe_o}
y(x) =  \sum_{i=1}^{n} G(x)_i \cdot E_i(x),
\end{equation}
where $G(x)_i$ stands for the probability of selecting expert $i$, and $\sum_i^n G(x)_i=1$. The expert network $E_i(x)$ is typically implemented by a few convolutional layers \cite{wang2020long} or multi-layer perceptions (MLP) layers \cite{chen2023adamv}. In this paper, we choose lightweight convolutional layers to restrict computational costs. In the MoE architecture, different experts share the same architecture but have isolated weights. The distribution among all $n$ experts is produced by the gating network $G$. Assuming the trainable weight matrix is $W_g$, the gating network $G$ is formulated as Eq. (\ref{eq:gate}).
\begin{equation}
\label{eq:gate}
G(x) = softmax(x\cdot W_g).
\end{equation}
The probability $G(x)_i$ is the $i$-th value of $G(x)$. To reduce the inference cost, current researches usually employ the sparse MoE \cite{shazeer2017outrageously} to select top-$k$ experts with the highest probabilities $G(x)_i$ as in Eq. (\ref{eq:smoe_o}).
\begin{equation}
\label{eq:smoe_o}
y(x) =  \sum_{i=1}^{k} G(x)_i \cdot E_i(x),
\end{equation}
where
\begin{equation}
\label{eq:smoe_p}
G(x) = TopK(softmax(x\cdot W_g),k).
\end{equation}
Here, $k$ is much smaller than the number of experts $n$.

\begin{figure}[t]
    \centering
    \includegraphics[width=1.0\linewidth]{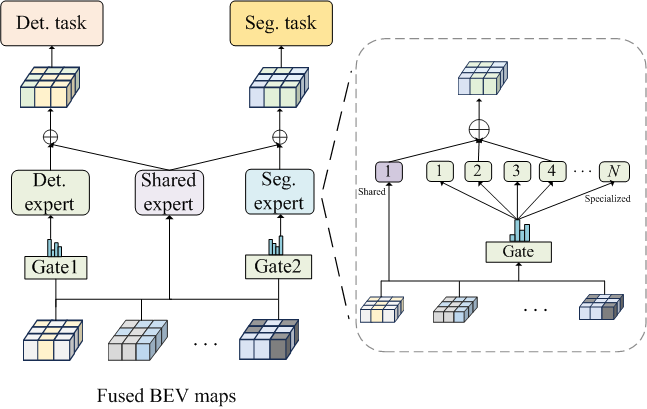}
    \caption{Illustration of the task-specific MoE module. The experts are split into detection experts, segmentation experts and shared experts. For each task, the corresponding specialized experts and shard ones are utilized to learn task-specific and task-invariant knowledge separately.}
    \label{fig:moe}
\end{figure}

The proposed task-specific MoE module is illustrated in Fig. \ref{fig:moe}. In joint learning for 3D object detection and map segmentation, there are 3 groups of experts, i.e., detection experts $E^{det}(x)$, segmentation experts $E^{seg}(x)$ and shared experts $\bar E(x)$. The detection experts focus on capturing spatial information and object relationships within the BEV features, while the segmentation experts emphasize region boundaries and pixel-level semantic classification. The shared experts are isolated from task-specific ones and are constantly active, allowing them to learn task-invariant knowledge from BEV features and enhance feature complementarity across tasks. To select sparse experts, two gates, i.e., $\widetilde G^{det}(x)$ and $\widetilde G^{seg}(x)$, are designed for 3D object detection and map segmentation separately. Such a design supports both single-task training by activating a specific gate and multi-task joint training by employing two separate gates for the tasks. Therefore, the task-specific MoEs for 3D object detection and map segmentation are separately formulated as
\begin{equation}
\label{eq:detmoe}
   y_{det}(x) = \sum_{i=1}^{k} \widetilde G^{det}(x)_i \cdot \widetilde E_i^{det}(x) + \frac{1}{s}\sum_{j=1}^{s} \bar E_j(x),
\end{equation}
and
\begin{equation}
\label{eq:segmoe}
   y_{seg}(x) = \sum_{i=1}^{k} \widetilde G^{seg}(x)_i \cdot \widetilde E_i^{seg}(x) + \frac{1}{s}\sum_{j=1}^{s} \bar E_j(x),
\end{equation} 
where the first items of both equations represent the specialized experts for the corresponding tasks, while the second items represent the shared experts. It is noted that all $s$ shared experts are assigned by even probabilities.

To alleviate the imbalanced expert load issue where some experts are overloaded while others are underutilized, we employ the noisy $TopK$ gating network \cite{shazeer2017outrageously} to avoid the overconfidence of the gating network at the beginning of training. 
Thus, the gating network in Eq. (\ref{eq:detmoe}) and Eq. (\ref{eq:segmoe}) is represented as follows.
\begin{equation}
\label{eq:topkn}
\small
\widetilde G(x) = TopK\left(Softmax( x \cdot W_g + \phi \cdot \log(1+e^x)(x\cdot W_{n})), k\right),
\end{equation}
where $\phi \sim \mathcal{N}(0, 1)$ represents a random variable sampled from a Gaussian distribution with zero mean and unit variance. The amount of noise is controlled by the trainable weight matrix $W_{n}$.

\subsection{Task heads and Losses}

\label{loss}

We employ task-specific heads for different tasks following the baselines \cite{liu2023bevfusion, wang2023unitr} used in this paper. For 3D object detection, the two-stage TransFusionHead \cite{bai2022transfusion} is adopted to obtain 3D detection boxes. In the first stage, a heatmap-based convolutional head combined with local Non-Maximum Suppression (NMS) generates initial object proposals, which are then selected as query vectors. In the second stage, a Transformer decoder iteratively refines these proposals using an attention mechanism, predicting offsets to update proposal positions, and thereby estimating object size and location. For semantic segmentation, a lightweight head is utilized. Specifically, feature maps are aligned with segmentation labels through \text{grid\_sample}, followed by a multi-layer convolutional classifier that processes semantic features of different categories. This produces class-specific probability maps, from which the final segmentation results are obtained.

This paper formulates the multi-task loss function as a combination of the 3D detection loss $L_{det3d}$ and the BEV segmentation loss $L_{seg}$.
\begin{equation}
\label{eq:loss}
L = \alpha\ L_{det3d} + \beta\ L_{seg},
\end{equation}
where $\alpha$ and $\beta$ are the weight coefficients of 3D object detection and map segmentation to balance the contribution of different tasks to the overall loss. Following the baselines \cite{liu2023bevfusion, wang2023unitr}, the weights $\alpha$ and $\beta$ are set to $1.0$ and $2.0$ respectively. The 3D detection loss $L_{det3d}$ includes a focal loss for classification, an $L_1$ Loss for bounding box regression, and a Gaussian focal loss for heatmap prediction, which is represented as follows.
\begin{equation}
\label{eq:det3d}
L_{det3d} = \frac{1}{N_{pos}} \left( \alpha_{cls} L_{cls} + \alpha_{heatmap} L_{heatmap} + \alpha_{bbox} L_{bbox} \right),
\end{equation}
where $N_{pos}$ is the number of positive samples. The values of $\alpha_{cls}$, $ \alpha_{heatmap}$ and $\alpha_{bbox}$ are set to $1.0$, $1.0$ and $0.25$ to make a fair comparison with the baselines. The standard focal loss \cite{lin2017focal} is utilized as the segmentation loss $L_{seg}$.

\section{Experiments}

\label{experiments}

\begin{table*}[tp]
    \centering
    \caption {\textbf{Comparisons on nuScenes validation set with the latest SOTA models. The symbols``C'' and ``L'' denote the camera and Lidar modalities, respectively. ``C+L'' indicates that the models are trained by inputting both the camera and Lidar data. ``Single-task learning'' refers to models in which 3D detection and map segmentation are trained independently. The joint-training models are denoted as ``multi-task learning''. The superscript ``$\ast$'' indicates the models are retrained under the same setting with MATS to ensure a fair comparison. The metrics mAP and NDS are employed for 3D detection, while the performance of the segmentation areas is quantified using mIoU.}}
    
    \label{tab:main}
    \resizebox{\textwidth}{!}{ 
    \renewcommand{\arraystretch}{1.1}
    \begin{tabular}{c|c|c|c|c|c|c|c|c|c|c} 
        \hline
        \toprule  
        \textbf{Methods} & \textbf{Modality} & \textbf{mAP} & \textbf{NDS} & \textbf{Drivable} & \textbf{Ped. Cross} & \textbf{Walkway} & \textbf{StopLine} & \textbf{Carpark}& \textbf{Divider} & \textbf{Mean} \\
        \hline  
        \textbf{Single-task learning} \\
        \hline
        BEVFusion\cite{liu2023bevfusion} & C & 35.6 & 41.2 & 81.7 & 54.8 & 58.4 & 47.4 & 50.7 & 46.4 & 56.6 \\
        X-Align\cite{borse2023x} & C & - & - & 82.4 & 55.6 & 59.3 & 49.6 & 53.8 & 47.4 & 58.0 \\
        M$^{2}$BEV\cite{xie2022m} & C & 41.7 & 47.0 & 77.2 & - & - & - & - & 40.5 & - \\
        BEVFormer\cite{li2024bevformer}   & C & 41.6 & 51.7 & 80.1 & - & - & - & - & 25.7 & - \\
        PointPillars\cite{lang2019pointpillars}  & L & 52.3 & 61.3 & 72.0 & 43.1 & 53.1 & 29.7 & 27.7 & 37.5 & 43.8 \\
        BEVFusion \cite{liu2023bevfusion}  & L & 64.7 & 69.3 & 75.6 & 48.4 & 57.5 & 36.4 & 31.7 & 41.9 & 48.6 \\
        CenterPoint\cite{yin2021center}  & L & 59.6 & 66.8 & 75.6 & 48.4 & 57.5 & 36.5 & 31.7 & 41.9 & 48.6 \\
        FocalFormer3D\cite{chen2023focalformer3d}  & L & 66.4 & 70.9 & - & - & - & - & - & - & - \\
        SAFDNet\cite{zhang2024safdnet}   & L & 66.1 & 70.0 & 76.1 & 48.7 & 57.0 & 36.9 & 33.0 & 42.2 & 49.0 \\

        X-Align\cite{borse2023x} & L+C & - & - & 86.8 & 65.2 & 70.0 & 58.3 & 57.1 & 58.2 & 65.7 \\
        TransFusion\cite{bai2022transfusion} & L+C & 67.3 & 71.2 & - & - & - & - & - & - & - \\
        PointPainting \cite{vora2020pointpainting}  & L+C & 65.8 & 69.6 & 75.9 & 48.5 & 57.1 & 36.9 & 34.5 & 41.9 & 49.1 \\

        FuTR3D \cite{chen2023futr3d}  & L+C & 64.5 & 68.3 & - & - & - & - & - & - & - \\
        MVP\cite{yin2021multimodal} & L+C & 66.1 & 70.0 & - & - & - & - & - & - & - \\
        MBFusion\cite{hao2024mbfusion}& L+C & - & - & - & - & - & - & - & - & 66.1\\
        BEVFusion \cite{liu2023bevfusion}$^\ast$ & L+C & 66.5 & 70.2 & 84.8 & 58.9 & 66.6 & 49.6 & 57.9 & 53.3 & 61.5 \\
        MapTR\cite{liao2022maptr} & L+C & - & - & - & 55.9 & - & - & - & - & 62.5 \\
        GraphBEV\cite{song2024graphbev} & L+C & - & - &  86.3& 60.9 & 69.1 & 53.1 & 57.5 & 53.1& 63.3 \\
        SMAB\cite{mustajbasic2025smab} & L+C & - & - & -& - & - & - & - & -& 63.4 \\
        CMGFA\cite{kuang2024cmgfa} & L+C & - & - &  87.5& 61.9 & 66.3 & 54.8 & 59.4 & 54.3& 64.0 \\
        \hline
        \textbf{Multi-task learning} \\
        \hline
        HENet++ \cite{xia2025henet++} & C & {56.7} & {63.7} & {-} & {-} & {-} & {-} & {-} & {-} & {58.3}\\
        {UniSparseBEV \cite{zhou2026unisparsebev}}& {C} & {45.1} & {55.4} & {-} & {-} & {-} & {-} & {-} & {-} & {49.3}\\
        {DAOcc\cite{yang2025daocc}$^\ast$} & {L+C} & {57.2} & {61.8} & {-} & {-} & {-} & {-} & {-} & {-} & {51.3}\\
        BEVFusion \cite{liu2023bevfusion}$^\ast $& L+C & 65.8 & 69.6 & 75.9 & 32.7 & 47.5& 20.3 & 32.8 & 31.7 & 40.4 \\
        FuLLER \cite{huang2023fuller}  & L+C & 60.5 & 65.3 & - & -& -& - & - & - & 58.4\\
        UniTR\cite{wang2023unitr}$^\ast$\  & L+C & 66.1 & 70.1 & 84.6 & 53.8& 62.7& 48.7 & 54.9 & 48.9 & 59.8\\
        
        MATS  & L+C & \textbf{67.5} & \textbf{70.8} & \textbf{86.2} & \textbf{61.9}& \textbf{69.3} & \textbf{54.7} & \textbf{59.2} & \textbf{55.3} & \textbf{64.6} \\
        \bottomrule  
    \end{tabular} 
    }
\end{table*}


In this section, we conduct comprehensive experiments to evaluate the proposed MATS. Specifically, we assess the performance of MATS on the public large-scale benchmark nuScene \cite{caesar2020nuscenes}, and compare it with the state-of-the-art (SOTA) to validate its effectiveness and superiority. Additionally, a robustness analysis and an ablation study are conducted to verify its flexibility and efficacy.

\subsection{Experimental setup}
\noindent\textbf{Dataset}. We evaluate MATS on the nuScenes dataset \cite{caesar2020nuscenes}. NuScenes contains $28,130$ training samples and $6,019$ validation
samples, each providing point clouds collected with a 32-beam LiDAR and images captured by 6-view cameras. For 3D object detection, 10 foreground categories are involved to assess the performance. This paper utilizes mean Average Precision (mAP) and nuScenes Detection Score (NDS) as the detection metrics by aligning with SOTAs \cite{liu2023bevfusion, wang2023unitr}. For map segmentation, the mean Intersection over Union (mIoU) is employed to assess the segmentation results on the overall six categories, including drivable
space, pedestrian crossing, walkway, stop line, car-parking area, and lane divider. In this paper, all models are trained on the training set and evaluated on the validation set.

\noindent\textbf{Implementation Details.} This paper adopts the multi-modal Transformer encoder in UniTR \cite{wang2023unitr} to generate the BEV feature maps from different modalities. The number of fusion branches in the modality-adaptive BEV fusion module is set to $N=3$. For each task, the specialized experts adopt $n=4$, where $k=2$ experts with the highest probabilities are selected to build the sparse MoE. The number of shared experts is set to $s=1$. 

Our model is trained on 4 GeForce RTX 4090D GPUs with $batch\_size=12$. AdamW optimizer \cite{loshchilov2017decoupled} with the initial learning rate $3 \times 10^{-3}$ is employed to update the weights of the network. The one-cycle learning policy \cite{smith2019super} is adopted to change the learning rate for all $20$ epochs.

\subsection{Main Results}
To verify the effectiveness and superiority of MATS, we mainly compare it with SOTA multi-task learning methods, including BEVFusion \cite{liu2023bevfusion}, FULLER \cite{huang2023fuller}, UniTR \cite{wang2023unitr}, HENet++ \cite{xia2025henet++}, UniSparseBEV \cite{zhou2026unisparsebev}, and DAOcc\cite{yang2025daocc}. Due to the limitation of GPU resources, the training batch size of the proposed network is much less than the SOTA methods. Therefore, we retrain BEVFusion and UniTR using their official codes by setting $batch\_size =12$ for a fair comparison. The experimental results are illustrated in TABLE \ref{tab:main}. ``C'' and ``L'' represent the modality of the camera and Lidar, respectively. ``C+L'' means that the models are trained by inputting both the camera and Lidar data. Single-task learning stands for that the models for 3D detection and map segmentation are trained in an isolated way. The joint-training models are represented by ``multi-task learning''. The symbol ``$\ast$'' indicates the corresponding models are retrained under the same setting with MATS for a fair comparison. Since FULLER \cite{huang2023fuller} is not open source, we do not retrain it in this paper. In TABLE \ref{tab:main}, the compared methods include single-task learning methods BEVFusion \cite{liu2023bevfusion}, X-Align\cite{borse2023x}, M$^{2}$BEV\cite{xie2022m}, BEVFormer\cite{li2024bevformer}, PointPillars\cite{lang2019pointpillars}, CenterPoint\cite{yin2021center}, FocalFormer3D\cite{chen2023focalformer3d}, SAFDNet\cite{zhang2024safdnet}, X-Align\cite{borse2023x}, TransFusion\cite{bai2022transfusion}, PointPainting \cite{vora2020pointpainting}, FuTR3D \cite{chen2023futr3d}, MVP\cite{yin2021multimodal}, MBFusion\cite{hao2024mbfusion}, MapTR\cite{liao2022maptr}, GraphBEV\cite{song2024graphbev}, SMAB\cite{mustajbasic2025smab}, CMGFA\cite{kuang2024cmgfa}, and multi-task learning methods BEVFusion \cite{liu2023bevfusion}, FuLLER \cite{huang2023fuller}, UniTR\cite{wang2023unitr}, HENet++ \cite{xia2025henet++}, UniSparseBEV \cite{zhou2026unisparsebev}, and DAOcc\cite{yang2025daocc}. From Table \ref{tab:main}, it is observed that there is an obvious negative transfer problem, where the SOTA multi-task learning methods are beaten by the single-task learning ones like X-Align \cite{borse2023x} and CMGFA \cite{kuang2024cmgfa}, especially on the map segmentation task. The proposed MATS significantly alleviates the problem by decoupling the tasks in joint learning. In comparison with the SOTA multi-task learning methods, MATS achieves the best performance on multi-task learning with $67.5\%$ mAP and $64.6\%$ mIoU. Compared to UniTR with the same encoders, MATS obtains $1.4\%$ mAP and $4.8\%$ mIoU gains, illustrating its effectiveness and superiority in 3D perception.

From Table \ref{tab:main}, it is also observed that the segmentation task achieves a larger performance gain than the detection task. This is primarily attributed to the fact that the proposed method enhances the gradient ratio of the segmentation task during back-propagation. Since the weights of the shared layers are updated according to the sum of the gradients generated by different tasks, the segmentation task with low gradient magnitudes in the baseline tends to be dominated by the detection task with larger gradient magnitudes. In the proposed framework, the designed modality-adaptive BEV fusion and task-specific MoE modules significantly increase the effective gradient ratio of the segmentation task, thereby directing more representational capability and optimization focus of the network toward segmentation.

\begin{figure}[htbp]
    \centering
    \includegraphics[width=0.85\columnwidth]{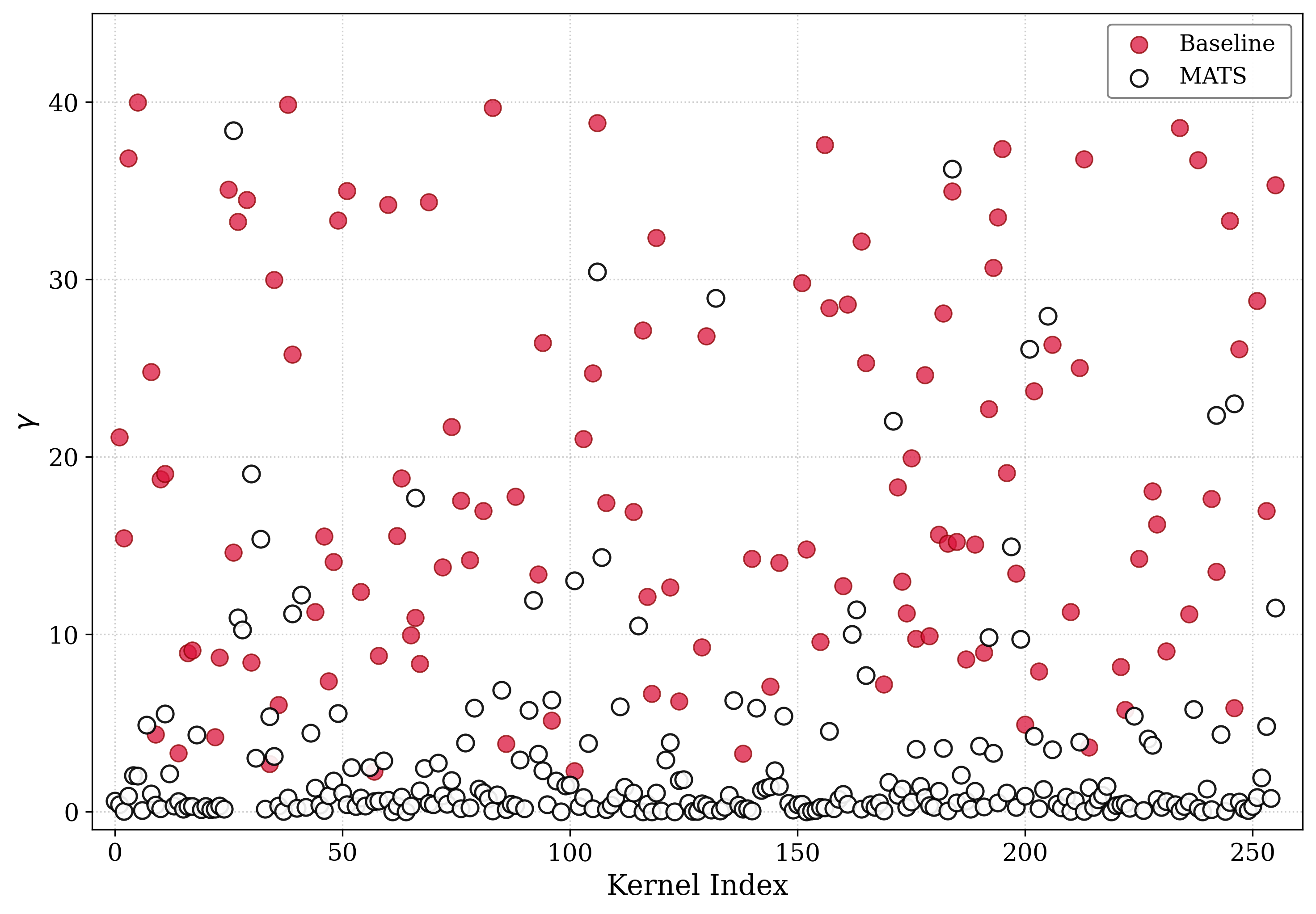}
    \vspace{0.5cm}
    \caption{We visualize the $\gamma$ values computed on the last shared layer preceding the decoder block. It is observed that the gradient magnitude associated with the segmentation loss is dominated by that of the detection loss in the baseline. In contrast, MATS effectively balances gradients arising from the two tasks, resulting in a more stable and balanced distribution of $\gamma$.}
    \label{fig:kernel_gamma_scatter}
\end{figure}

To validate the above analysis, we visualize the distribution of the gradient ratios between the detection and the segmentation tasks in the last shared layer, as shown in Fig. \ref{fig:kernel_gamma_scatter}. Specifically, the gradient ratio $\gamma$ is calculated as Eq. (\ref{eq:gamma_ratio}).


\begin{equation}
\gamma = \frac{\left\| \nabla_{\text{shared\_L}} L_{det3d} \right\|}{\left\| \nabla_{\text{shared\_L}} L_{seg} \right\|}, 
\label{eq:gamma_ratio}
\end{equation}
where the numerator stands for the $L2$ norm of the gradient corresponding to the objection task, while the denominator denotes the $L2$ norm for the segmentation task. Since there are 256 kernels in the last shared layer, Fig. \ref{fig:kernel_gamma_scatter} shows a scatter plot of the gradient ratios of the baseline and the proposed MATS. As illustrated in Fig. \ref{fig:kernel_gamma_scatter}, the proposed method effectively balances the gradients from the two tasks, leading to a more stable and lower gradient ratio $\gamma$. By balancing the gradient ratio, the performance of the segmentation task is significantly improved.


\begin{table}[t]
    \centering
    \caption{Comparison with SOTAs using the same BEV encoders on nuScenes validation set.}
    \label{tab:backbone}
    \renewcommand{\arraystretch}{1.1}
    \hspace{-0.1cm}
    \begin{tabular}{c@{\hspace{2pt}}|c@{\hspace{1pt}}|c@{\hspace{3pt}}|c@{\hspace{15pt}}c@{\hspace{8pt}}c@{\hspace{8pt}}c@{\hspace{3pt}}c}
        \hline
        \textbf{\multirow{2}{*}{Task type}} & \textbf{\multirow{2}{*}{Method}} &\textbf{\multirow{2}{1.2cm}{Param(M)}} & \multicolumn{2}{c}{\textbf{Det.}}& &\textbf{Seg.}\\
        \cline{4-5}\cline{7-7}
        &&&\textbf{ mAP}&\textbf{NDS}&&\textbf{mIoU}\\
        \hline
         \multirow{8}{*}{\shortstack{Single-task\\ learning}} & \multirow{2}{*}{BEVFusion}& 40.9M & 66.5 & 70.2 & &- \\
        && 43.1M  & - & -&& 61.5\\
      
        & \multirow{2}{*}{BEVFu.+MATS} & 45.3M & 66.9& 70.3 & &- \\
        &&  47.6M & -& -&& 65.2 \\
        \cline{2-7}
         & \multirow{2}{*}{UniTR}& 37.5M & 66.7 & 70.3 & &- \\
        &&  38.2M & - & - && 63.2\\
    
        & \multirow{2}{*}{UniTR+MATS} & 42.3M & 67.9& 71.1 & &- \\
        && 43.7M   & -& -&&66.5\\
       \hline

        \multirow{4}{*}{\shortstack{Multi-task \\ learning}} & BEVFusion & 45.6M & 65.8 & 69.6 && 40.4 \\ 

        &BEVFu.+MATS& 51.3M & 66.8 & 70.3 && 50.2\\
       
        \cline{2-7}
        & UniTR & 39.3M & 66.1 & 70.1&& 59.8 \\

        & UniTR+MATS& 44.5M     & 67.5 & 70.8 && 64.6\\
      \hline
    \end{tabular}
\end{table}

\subsection{Robustness analysis}

 This paper further evaluates MATS on different BEV encoders and task types to validate its robustness. Specifically, the BEV encoders from BEVFusion \cite{liu2023bevfusion} and UniTR \cite{wang2023unitr} are employed to test the performance of MATS on both the joint task and the separate tasks. The evaluation results are shown in TABLE \ref{tab:backbone}. We can see from TABLE \ref{tab:backbone} that MATS brings consistent perception improvements on different BEV encoders and task types. In multi-task learning, MATS with the BEVFusion encoder obtains a $1\%$ mAP gain and a $0.7\%$ NDS gain on 3D detection and a $9.8\%$ mIoU gain on segmentation. Using the UniTR encoder, MATS achieves performance gains of $0.6\%$ in mAP, $0.7\%$ in NDS and $4.8\%$ in mIoU. Meanwhile, the same trend of performance improvement is shown in single-task learning, demonstrating the robustness of MATS against BEV encoders and task types. TABLE \ref{tab:backbone} also displays the number of parameters of different models. Compared to single-task learning models, the proposed MATS significantly mitigates the negative transfer problem with a limited increase in parameter number, where MATS only endures a drop of $1.9\%$ mIoU with the UniTR encoder, while the baseline with the same encoder suffers from a reduction of $3.4\%$ in mIoU.

\subsection{Ablation studies and Analysis}
This section aims to investigate the contribution of each proposed component to the performance improvement and the effects of the hyperparameters in MATS.

\subsubsection{Component analysis}
We conducted an ablation study on the proposed modality-adaptive fusion module and the task-specific MoE module. The proposed modules are successively added to the baseline (UniTR\cite{wang2023unitr}) to evaluate their contributions. The evaluation results are shown in TABLE \ref{tab:ablation}. The first row represents the results achieved by the baseline model. From TABLE \ref{tab:ablation}, it is observed that the largest performance improvement is achieved by the modality-adaptive fusion module, where $1.1\%$ mAP, $0.5\%$ NDS and $3.7\%$ mIoU gains are obtained, verifying the analysis in Section \ref{intro} about the weakness of the single BEV feature map. When further inserting the MoE module into the baseline, the segmentation performance is significantly improved, reaching $64.6\%$ mIoU. 

\begin{table}[t]
\centering
\caption{\textbf{Ablation study on the proposed modality-adaptive BEV fusion module and task-
specific MoE module.}}
\renewcommand{\arraystretch}{1.1}
\begin{tabular}{c@{\hspace{3pt}}c@{\hspace{20pt}}c@{\hspace{28pt}}cc@{\hspace{17pt}}c}  
       \hline  
    \textbf{\multirow{2}{*}{\shortstack{Modality-adaptive\\ BEV fusion}}} 
      & \textbf{\multirow{2}{*}{\shortstack{Task-specific\\ MoE}}} & \multicolumn{2}{c}{\textbf{Det.}}&& \textbf{Seg.} \\
    \cline{3-4} \cline{6-6}
      & & \textbf{map}& \textbf{NDS}& &\textbf{mIoU}\\
    \hline
        \ - & -  & 66.1 & 70.1 && 59.8 \\
        \checkmark & - & 67.2 & 70.6& & 63.5 \\
        \ - &\checkmark  & 66.9 & 70.3& & 61.3 \\
        \checkmark & \checkmark &  \textbf{67.5} & \textbf{70.8} && \textbf{64.6} \\
        \hline  
    \end{tabular}  
    \label{tab:ablation}
\end{table}

\subsubsection{The effect of the number of the fused feature maps}
The number of fused feature maps $N$ is an important hyperparameter. A limited number of fusion maps carry insufficient complementary information from different modalities, while excessive fusion maps are likely to introduce redundant information for downstream networks and increase the computation cost. TABLE \ref{tab:num_fusion} tests different numbers of the fusion maps with a range of $\left [1,5\right ]$ and reports their performance on nuScene validation set. For TABLE \ref{tab:num_fusion}, it is evident that the highest mAP, NDS and mIoU are achieved when $N=3$. As the value of $N$ decreases, the complementary information from different modalities cannot be fully carried out by the fusion maps. In contrast, increasing the value of $N$ results in redundancy in information for the following task-specific MoE module, interfering with the acquisition of the most related features from the fusion maps for the tasks.

\begin{table}[htbp]
    \centering  
    \caption{The experimental results under the different numbers of the fused BEV feature maps in the modality-adaptive fusion module.}  
    \begin{tabular}{c@{\hspace{30pt}}c@{\hspace{30pt}}c@{\hspace{6pt}}c@{\hspace{25pt}}c}  
        \hline
         \textbf{\multirow{2}{*}{$N$}} & \multicolumn{2}{c}{\textbf{Det.}}&&\textbf{Seg}\\
         \cline{2-3}\cline{5-5}
         &\textbf{mAP}&\textbf{NDS} &&\textbf{mIoU} \\
        \hline
         1 & 66.3 & 70.2  && 60.5 \\
         2 & 66.9 &  70.4 && 63.5 \\
         3 &  \textbf{67.5}& \textbf{70.8} &&\textbf{64.6}\\ 
         4 &67.1 & 70.6 && 64.1   \\
         5 &  67.0&   70.5&&  63.9\\
        \hline
    \end{tabular}  
    \label{tab:num_fusion}  
\end{table}

\subsubsection{The effect of the number of the shared experts} 
\begin{figure}[t]
    \centering
    \includegraphics[width=1.0\linewidth]{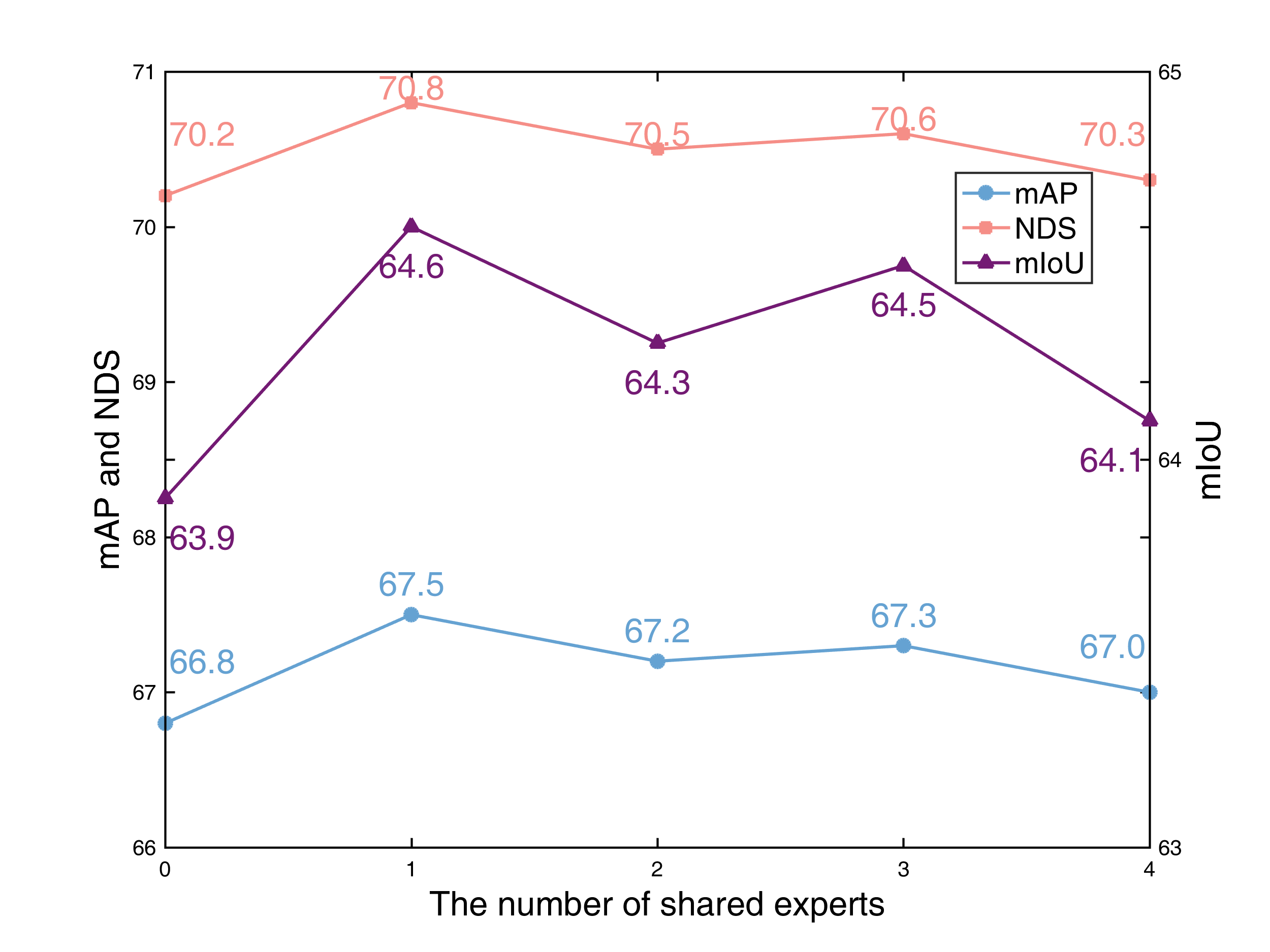}
    \caption{The experimental results along with the increase of the number of shared experts in the task-specific MoE module.}
    \label{fig:shared}
\end{figure}

To verify the contribution of the shared experts in MoE, this section examines the performance of MATS under different numbers of shared experts. The performance evaluated on nuScenes validation set is illustrated in Fig. \ref{fig:shared}. The x-axis represents the number of the shared experts $s$. The left y-axis marks the mAP and NDS values for 3D detection, while mIoU is calibrated in the right y-axis. From Fig. \ref{fig:shared}, it is obvious that the model trained by $1$ shared expert achieves the best performance. The model without shared experts ($s=0$) cannot provide common knowledge for different tasks, whereas excessively shared experts tend to hijack the attention of the tasks from the focus knowledge.

\subsubsection{The performance comparison under different architectural designs for the fusion module}
To evaluate the rationality of the proposed multi-branch fusion architecture in the modality-adaptive BEV fusion module, we compare it against two alternative designs: Single-Branch Multi-Channel (SB-MC) fusion and Shared-Branch Multi-Head (SB-MH) fusion. The SB-MC fusion generates a unified feature map whose channel dimension is $3\times$ that of the fused feature map obtained by the proposed multi-branch fusion. The unified feature map is evenly split into three groups for downstream tasks. In the SB-MH fusion, the shared branch generates three identical feature maps. These identical feature maps are fed into different expert networks for multi-task learning. The assessment results are shown in TABLE \ref{tab:fusion_module_table}. Compared with the proposed architecture, both the SB-MC fusion and the SB-MH fusion require all tasks to share a common fusion branch, resulting in gradient direction conflicts during backpropagation. In contrast, the multi-branch architecture design effectively mitigates mutual interference among different tasks and achieves the best performance when compared with the two alternative designs.


\begin{table}[htbp]
    \centering
    \caption{The performance comparison under different architectural designs for the fusion module.}
    \label{tab:fusion_module_table}
    \begin{tabular}{l@{\hspace{30pt}}c@{\hspace{30pt}}c@{\hspace{6pt}}c@{\hspace{25pt}}c}
        \toprule
        \textbf{\multirow{2}{*}{Fusion Module}} & \multicolumn{2}{c}{\textbf{Det.}} & & \textbf{Seg.} \\
        \cmidrule{2-3} \cmidrule{5-5}
        & \textbf{mAP} & \textbf{NDS} & & \textbf{mIoU} \\
        \midrule
        SB-MC Fusion & 66.2 & 70.1 & & 62.1 \\
        SB-MH Fusion & 66.8 & 70.4& & 62.6  \\
        Multi-Branch Fusion(ours) & \textbf{67.5} & \textbf{70.8} & & \textbf{64.6} \\
        \bottomrule
    \end{tabular}
\end{table}

\subsection{The generalization ability of the proposed method}

The generalization ability of the proposed multi-task learning framework is further evaluated by extending it to more tasks. In this section, we conduct multi-task joint training for 3D object detection, 3D occupancy prediction, and BEV map segmentation. To implement the experiment, DAOcc \cite{yang2025daocc} is selected as the baseline. The proposed modality-adaptive BEV fusion module and task-specific MoE module are inserted into DAOcc for a fair comparison. All experiments adopt RetNet-50 as the backbone network, with LiDAR and camera modalities as model inputs. TABLE \ref{tab:occ_comparison} reports the quantitative results on both two-task and three-task joint training configurations. From TABLE \ref{tab:occ_comparison}, it is obvious that the proposed method significantly alleviates the negative transfer effect. In the original DAOcc, two-task learning leads to performance drops of 2.8$\%$ mAP and 1.1$\%$ mIoU for detection and occupancy, respectively. Similarly, a noticeable accuracy degradation is also observed under three-task learning setting. After incorporating the proposed modality-adaptive BEV fusion module and task-specific MoE module, the detection mAP and the occupancy mIoU in two-task learning increase from 62.3$\%$ to 63.9$\%$ and from 46.8$\%$ to 49.2$\%$, respectively. For three-task joint training, the detection mAP, occupancy mIoU and segmentation mIoU are improved by 1.9$\%$, 2.9$\%$ and 2.4$\%$, respectively. These results indicate that the proposed method effectively mitigates the negative transfer issue across diverse tasks and thereby demonstrates strong generalization ability in the multi-task learning setting.

\begin{table}[t]
\centering  
\caption{The experimental results on nuScenes for the 3D Object Detection, 3D Occupancy Prediction, and BEV Segmentation tasks. Both two-task and three-task joint training settings are included in the experiment. Best results are highlighted in \textbf{bold}.}
\label{tab:occ_comparison}
\resizebox{\linewidth}{!}{%
\begin{tabular}{lccccccc}
        \hline 
        \textbf{\multirow{2}{*}{\centering Method}} &
        \multicolumn{2}{c}{\multirow{1}{*}{\textbf{Det.}}} & &
        \multicolumn{1}{c}{\multirow{1}{*}{\textbf{Occ.}}} &
        &\multicolumn{1}{c}{\multirow{1}{*}{\textbf{Seg.}}}\\
        \cline{2-3} \cline{5-5}  \cline{7-7}
         & \textbf{\multirow{1}{*}{mAP}} & \textbf{NDS} & &\textbf{mIoU} &
    &\textbf{mIoU} \\
   \hline 
   Detection and Occpuancy tasks\\
   \hline
DAOcc(STL)        & 62.3 & 66.1 &&-&&-\\
DAOcc(STL)        & - & - & &47.9 &&-\\
DAOcc(MTL)        & 59.5 & 63.8 & &46.8 &&- \\
DAOcc+MATS(MTL)  & \textbf{61.6} & \textbf{65.7} & &\textbf{49.2}&& -\\
\hline
Detection, Occupancy, and BEV Segmentation tasks\\
\hline
DAOcc(MTL)        & 57.2 & 61.8 & &45.9 & &51.3\\
DAOcc+MATS(MTL)   & \textbf{59.1} & \textbf{63.0} & & \textbf{48.8}& &\textbf{53.7}\\
\hline 
\end{tabular}%
}
\end{table}

\begin{figure*}
    \centering
    \includegraphics[width=0.9\linewidth]{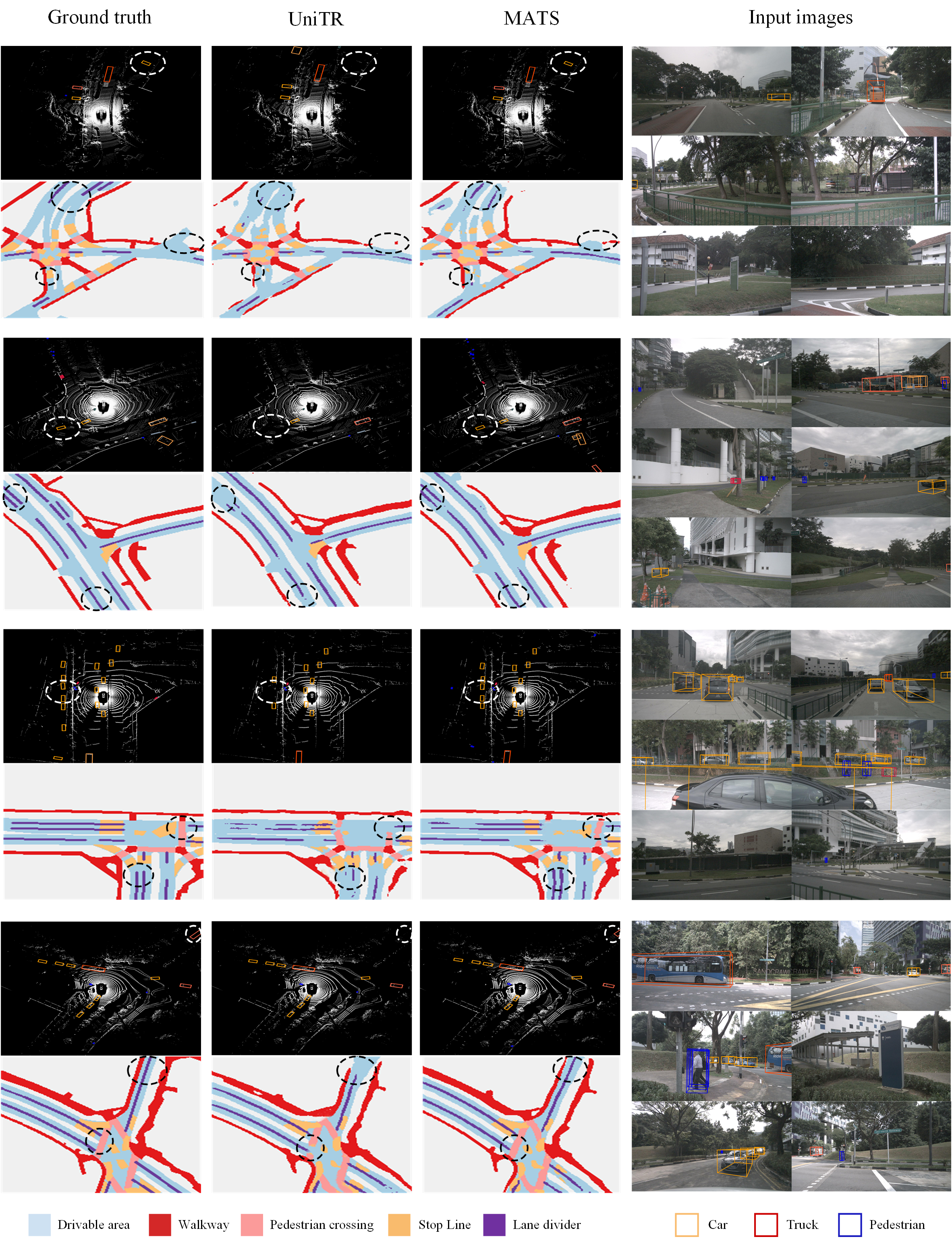}
    \vspace{0.5cm}
    \caption{Visualization results of the baseline UniTR and the proposed MATS on multi-task 3D perception, including 3D object detection and BEV map segmentation. MATS shows better results than UniTR. The improved areas are highlighted by dashed circles.}
    \label{fig:visualunitr}
\end{figure*}

\subsection{Visualization Results}

Fig. \ref{fig:visualunitr} illustrates the multi-task perception results of the baseline UniTR \cite{wang2023unitr} and MATS on nuScenes validation set. The first column represents the ground truth from nuScenes. The predicted results of UniTR and MATS are displayed in the second and third columns. The input images are also listed in the last column of Fig. \ref{fig:visualunitr} for an intuitive observation of the objects in the detection and segmentation tasks. From Fig. \ref{fig:visualunitr}, it is observed that MATS achieves better prediction results than UniTR. Taking the first-row results as an example, in the detection task, the baseline model UniTR fails to detect the car target highlighted in the white dash circle, whereas MATS successfully identifies it. In the segmentation task, UniTR exhibits poor performance in lane divider segmentation, while MATS not only improves the segmentation performance of lane divider (the top area highlighted by the black dashed circle) but also achieves more precise walkway segmentation results (the bottom-left area highlighted by the black dashed circle). Similarly, as shown in the third-row images, UniTR misses several cars on the left of the ego and part of the pedestrian crossing and lane divider regions, whereas MATS shows a strong ability to perceive the surrounding dynamic objects and static areas.

\section{Conclusion}
\label{conclusion}
This paper proposes a novel multi-modality multi-task learning approach for 3D perception in autonomous driving. Instead of designing intricate fusion strategies that compress multimodal information into a single BEV feature map with limited capacity, our approach adopts a simple fusion network that generates multiple fused BEV feature maps to better capture complementary cross-modal information. To mitigate task conflicts, a task-specific Mixture-of-Experts (MoE) module is introduced to decouple different tasks and enable the network to automatically select the most suitable fused BEV features for each. Experimental results on the large-scale nuScenes dataset demonstrate that the proposed method achieves state-of-the-art performance in 3D perception, highlighting its strong potential for real-world deployment in autonomous driving systems.

\bibliographystyle{IEEEtran}
\bibliography{ref}

\begin{thebibliography}{10}
\providecommand{\url}[1]{#1}
\csname url@samestyle\endcsname
\providecommand{\newblock}{\relax}
\providecommand{\bibinfo}[2]{#2}
\providecommand{\BIBentrySTDinterwordspacing}{\spaceskip=0pt\relax}
\providecommand{\BIBentryALTinterwordstretchfactor}{4}
\providecommand{\BIBentryALTinterwordspacing}{\spaceskip=\fontdimen2\font plus
\BIBentryALTinterwordstretchfactor\fontdimen3\font minus \fontdimen4\font\relax}
\providecommand{\BIBforeignlanguage}[2]{{%
\expandafter\ifx\csname l@#1\endcsname\relax
\typeout{** WARNING: IEEEtran.bst: No hyphenation pattern has been}%
\typeout{** loaded for the language `#1'. Using the pattern for}%
\typeout{** the default language instead.}%
\else
\language=\csname l@#1\endcsname
\fi
#2}}
\providecommand{\BIBdecl}{\relax}
\BIBdecl

\bibitem{alemaw2025modeling}
A.~S. Alemaw, G.~Slavic, P.~Zontone, L.~Marcenaro, D.~M. Gomez, and C.~Regazzoni, ``Modeling interactions between autonomous agents in a multi-agent self-awareness architecture,'' \emph{IEEE Transactions on Multimedia}, 2025.

\bibitem{wang2026privacy}
S.~Wang, L.~Li, M.~Santos, and G.~Wang, ``Privacy-concealing cooperative perception for bev scene segmentation,'' in \emph{2026 IEEE International Conference on Acoustics, Speech and Signal Processing (ICASSP)}.\hskip 1em plus 0.5em minus 0.4em\relax IEEE, 2026, pp. 22\,232--22\,236.

\bibitem{shi2024nitedr}
C.~Shi, L.~Fang, H.~Wu, X.~Xian, Y.~Shi, and L.~Lin, ``Nitedr: Nighttime image de-raining with cross-view sensor cooperative learning for dynamic driving scenes,'' \emph{IEEE Transactions on Multimedia}, vol.~26, pp. 9203--9215, 2024.

\bibitem{wang2025ubtransformer}
K.~Wang, Q.~Ma, X.~Li, C.~Shen, R.~Leng, and J.~Lu, ``Ubtransformer: Uncertainty-based transformer model for complex scenarios detection in autonomous driving,'' \emph{IEEE Transactions on Multimedia}, 2025.

\bibitem{guesmi2023physical}
A.~Guesmi, M.~A. Hanif, B.~Ouni, and M.~Shafique, ``Physical adversarial attacks for camera-based smart systems: Current trends, categorization, applications, research challenges, and future outlook,'' \emph{IEEE Access}, vol.~11, pp. 109\,617--109\,668, 2023.

\bibitem{zhao2023lif}
L.~Zhao, H.~Zhou, X.~Zhu, X.~Song, H.~Li, and W.~Tao, ``Lif-seg: Lidar and camera image fusion for 3d lidar semantic segmentation,'' \emph{IEEE Transactions on Multimedia}, vol.~26, pp. 1158--1168, 2023.

\bibitem{zhang2025synet}
X.~Zhang, K.~Bi, S.~Chan, S.~Lu, and X.~Zhou, ``Synet: A synergistic network for 3d object detection through geometric-semantic-based multi-interaction fusion,'' \emph{IEEE Transactions on Multimedia}, 2025.

\bibitem{xiang2023multi}
C.~Xiang, C.~Feng, X.~Xie, B.~Shi, H.~Lu, Y.~Lv, M.~Yang, and Z.~Niu, ``Multi-sensor fusion and cooperative perception for autonomous driving: A review,'' \emph{IEEE Intelligent Transportation Systems Magazine}, vol.~15, no.~5, pp. 36--58, 2023.

\bibitem{huo2026cgmae}
J.~Huo, S.~Wang, E.~Chen, Y.~Ding, and S.~Yang, ``Cg-mae: Bev masked autoencoders based on cross-modal guidance for 3d object detection in autonomous driving,'' \emph{Computers, Materials \& Continua}, pp. 1--21, 2026.

\bibitem{bai2022transfusion}
X.~Bai, Z.~Hu, X.~Zhu, Q.~Huang, Y.~Chen, H.~Fu, and C.-L. Tai, ``Transfusion: Robust lidar-camera fusion for 3d object detection with transformers,'' in \emph{Proceedings of the IEEE/CVF conference on computer vision and pattern recognition}, 2022, pp. 1090--1099.

\bibitem{hao2025mapfusion}
X.~Hao, Y.~Diao, M.~Wei, Y.~Yang, P.~Hao, R.~Yin, H.~Zhang, W.~Li, S.~Zhao, and Y.~Liu, ``Mapfusion: A novel bev feature fusion network for multi-modal map construction,'' \emph{Information Fusion}, vol. 119, p. 103018, 2025.

\bibitem{liu2023bevfusion}
Z.~Liu, H.~Tang, A.~Amini, X.~Yang, H.~Mao, D.~L. Rus, and S.~Han, ``Bevfusion: Multi-task multi-sensor fusion with unified bird's-eye view representation,'' in \emph{2023 IEEE international conference on robotics and automation (ICRA)}.\hskip 1em plus 0.5em minus 0.4em\relax IEEE, 2023, pp. 2774--2781.

\bibitem{ma2024mta}
Y.~Ma, B.~Yaman, X.~Ye, J.~Luo, F.~Tao, A.~Mallik, Z.~Wang, and L.~Ren, ``Mta: Multimodal task alignment for bev perception and captioning,'' in \emph{Proceedings of the IEEE/CVF International Conference on Computer Vision}, 2026, pp. 670--679.

\bibitem{chen2025m3net}
X.~Chen, S.~Shi, T.~Ma, J.~Zhou, S.~See, K.~C. Cheung, and H.~Li, ``M3net: Multimodal multi-task learning for 3d detection, segmentation, and occupancy prediction in autonomous driving,'' in \emph{Proceedings of the AAAI Conference on Artificial Intelligence}, vol.~39, no.~2, 2025, pp. 2275--2283.

\bibitem{huang2023fuller}
Z.~Huang, S.~Lin, G.~Liu, M.~Luo, C.~Ye, H.~Xu, X.~Chang, and X.~Liang, ``Fuller: Unified multi-modality multi-task 3d perception via multi-level gradient calibration,'' in \emph{Proceedings of the IEEE/CVF International Conference on Computer Vision}, 2023, pp. 3502--3511.

\bibitem{xie2022m}
E.~Xie, Z.~Yu, D.~Zhou, J.~Philion, A.~Anandkumar, S.~Fidler, P.~Luo, and J.~M. Alvarez, ``M2bev: multi-camera joint 3d detection and segmentation with unified birds-eye view representation,'' \emph{arXiv preprint arXiv:2204.05088}, 2022.

\bibitem{wang2023unitr}
H.~Wang, H.~Tang, S.~Shi, A.~Li, Z.~Li, B.~Schiele, and L.~Wang, ``Unitr: A unified and efficient multi-modal transformer for bird's-eye-view representation,'' in \emph{Proceedings of the IEEE/CVF international conference on computer vision}, 2023, pp. 6792--6802.

\bibitem{zhao2024maskbev}
X.~Zhao, X.~Zhang, D.~Yang, M.~Sun, M.~Li, S.~Wang, and L.~Zhang, ``Maskbev: Towards a unified framework for bev detection and map segmentation,'' in \emph{Proceedings of the 32nd ACM International Conference on Multimedia}, 2024, pp. 2652--2661.

\bibitem{caesar2020nuscenes}
H.~Caesar, V.~Bankiti, A.~H. Lang, S.~Vora, V.~E. Liong, Q.~Xu, A.~Krishnan, Y.~Pan, G.~Baldan, and O.~Beijbom, ``nuscenes: A multimodal dataset for autonomous driving,'' in \emph{Proceedings of the IEEE/CVF conference on computer vision and pattern recognition}, 2020, pp. 11\,621--11\,631.

\bibitem{guo2020deep}
Y.~Guo, H.~Wang, Q.~Hu, H.~Liu, L.~Liu, and M.~Bennamoun, ``Deep learning for 3d point clouds: A survey,'' \emph{IEEE transactions on pattern analysis and machine intelligence}, vol.~43, no.~12, pp. 4338--4364, 2020.

\bibitem{patole2017automotive}
S.~M. Patole, M.~Torlak, D.~Wang, and M.~Ali, ``Automotive radars: A review of signal processing techniques,'' \emph{IEEE Signal Processing Magazine}, vol.~34, no.~2, pp. 22--35, 2017.

\bibitem{agrawal2023static}
S.~Agrawal, S.~Bhanderi, K.~Doycheva, and G.~Elger, ``Static multitarget-based autocalibration of rgb cameras, 3-d radar, and 3-d lidar sensors,'' \emph{IEEE Sensors Journal}, vol.~23, no.~18, pp. 21\,493--21\,505, 2023.

\bibitem{wang2020high}
L.~Wang, T.~Chen, C.~Anklam, and B.~Goldluecke, ``High dimensional frustum pointnet for 3d object detection from camera, lidar, and radar,'' in \emph{2020 IEEE Intelligent Vehicles Symposium (IV)}.\hskip 1em plus 0.5em minus 0.4em\relax IEEE, 2020, pp. 1621--1628.

\bibitem{li2022mathsf}
Y.~Li, J.~Deng, Y.~Zhang, J.~Ji, H.~Li, and Y.~Zhang, ``Ezfusion: A close look at the integration of lidar, millimeter-wave radar, and camera for accurate 3d object detection and tracking,'' \emph{IEEE Robotics and Automation Letters}, vol.~7, no.~4, pp. 11\,182--11\,189, 2022.

\bibitem{ravindran2022camera}
R.~Ravindran, M.~J. Santora, and M.~M. Jamali, ``Camera, lidar, and radar sensor fusion based on bayesian neural network (clr-bnn),'' \emph{IEEE Sensors Journal}, vol.~22, no.~7, pp. 6964--6974, 2022.

\bibitem{Chu_2023_WACV}
S.-Y. Chu and M.-S. Lee, ``Mt-detr: Robust end-to-end multimodal detection with confidence fusion,'' in \emph{Proceedings of the IEEE/CVF Winter Conference on Applications of Computer Vision (WACV)}, January 2023, pp. 5252--5261.

\bibitem{kim2024rcm}
J.~Kim, M.~Seong, G.~Bang, D.~Kum, and J.~W. Choi, ``Rcm-fusion: Radar-camera multi-level fusion for 3d object detection,'' in \emph{2024 IEEE International Conference on Robotics and Automation (ICRA)}.\hskip 1em plus 0.5em minus 0.4em\relax IEEE, 2024, pp. 18\,236--18\,242.

\bibitem{zhou2023bridging}
T.~Zhou, J.~Chen, Y.~Shi, K.~Jiang, M.~Yang, and D.~Yang, ``Bridging the view disparity between radar and camera features for multi-modal fusion 3d object detection,'' \emph{IEEE Transactions on Intelligent Vehicles}, vol.~8, no.~2, pp. 1523--1535, 2023.

\bibitem{harley2022simple}
A.~W. Harley, Z.~Fang, J.~Li, R.~Ambrus, and K.~Fragkiadaki, ``Simple-bev: What really matters for multi-sensor bev perception?'' in \emph{2023 IEEE International Conference on Robotics and Automation (ICRA)}.\hskip 1em plus 0.5em minus 0.4em\relax IEEE, 2023, pp. 2759--2765.

\bibitem{hazarika2025multifusionnet}
A.~Hazarika, M.~Fotouhi, M.~Rahmati, P.~Arabshahi, and W.~Cheng, ``Multifusionnet: Spatio-temporal camera-radar fusion in dynamic urban environments,'' \emph{IEEE Sensors Journal}, 2025.

\bibitem{lin2024rcbevdet}
Z.~Lin, Z.~Liu, Z.~Xia, X.~Wang, Y.~Wang, S.~Qi, Y.~Dong, N.~Dong, L.~Zhang, and C.~Zhu, ``Rcbevdet: Radar-camera fusion in bird's eye view for 3d object detection,'' in \emph{Proceedings of the IEEE/CVF Conference on Computer Vision and Pattern Recognition}, 2024, pp. 14\,928--14\,937.

\bibitem{fu2024eliminating}
J.~Fu, C.~Gao, Z.~Wang, L.~Yang, X.~Wang, B.~Mu, and S.~Liu, ``Eliminating cross-modal conflicts in bev space for lidar-camera 3d object detection,'' in \emph{2024 IEEE International Conference on Robotics and Automation (ICRA)}.\hskip 1em plus 0.5em minus 0.4em\relax IEEE, 2024, pp. 16\,381--16\,387.

\bibitem{xia2024henet}
Z.~Xia, Z.~Lin, X.~Wang, Y.~Wang, Y.~Xing, S.~Qi, N.~Dong, and M.-H. Yang, ``Henet: Hybrid encoding for end-to-end multi-task 3d perception from multi-view cameras,'' in \emph{European Conference on Computer Vision}.\hskip 1em plus 0.5em minus 0.4em\relax Springer, 2024, pp. 376--392.

\bibitem{xiao2025adversarial}
X.~Xiao, Q.~V. Hu, T.~H. Kim, and G.~Wang, ``Adversarial multi-task learning for liver tumor segmentation, dynamic enhancement regression, and classification,'' \emph{arXiv preprint arXiv:2511.20793}, 2025.

\bibitem{li2024quadbev}
Y.~Li, Y.~Li, X.~Yang, M.~Yu, Z.~Huang, X.~Wu, and C.~Yeo, ``Quadbev: An efficient quadruple-task perception framework via birds'-eye-view representation,'' in \emph{2024 IEEE 27th International Conference on Intelligent Transportation Systems (ITSC)}.\hskip 1em plus 0.5em minus 0.4em\relax IEEE, 2024, pp. 2405--2412.

\bibitem{hao2025msc}
X.~Hao, G.~Liu, Y.~Zhao, Y.~Ji, M.~Wei, H.~Zhao, L.~Kong, R.~Yin, and Y.~Liu, ``Msc-bench: Benchmarking and analyzing multi-sensor corruption for driving perception,'' in \emph{2025 IEEE International Conference on Multimedia and Expo (ICME)}.\hskip 1em plus 0.5em minus 0.4em\relax IEEE, 2025, pp. 1--6.

\bibitem{yu2025sgformer}
H.~Yu, S.~Chan, X.~Zhou, and X.~Zhang, ``Sgformer: Semantic-geometry fusion transformer for multi-modal 3d panoptic segmentation,'' in \emph{Proceedings of the AAAI Conference on Artificial Intelligence}, vol.~39, no.~9, 2025, pp. 9616--9625.

\bibitem{song2024graphbev}
Z.~Song, L.~Yang, S.~Xu, L.~Liu, D.~Xu, C.~Jia, F.~Jia, and L.~Wang, ``Graphbev: Towards robust bev feature alignment for multi-modal 3d object detection,'' in \emph{European Conference on Computer Vision}.\hskip 1em plus 0.5em minus 0.4em\relax Springer, 2024, pp. 347--366.

\bibitem{kuang2024cmgfa}
X.~Kuang, R.~Niu, C.~Hua, C.~Jiang, H.~Zhu, Z.~Chen, and B.~Yu, ``Cmgfa: A bev segmentation model based on cross-modal group-mix attention feature aggregator,'' \emph{IEEE Robotics and Automation Letters}, 2024.

\bibitem{hsiao2019filter}
T.-Y. Hsiao, Y.-C. Chang, H.-H. Chou, and C.-T. Chiu, ``Filter-based deep-compression with global average pooling for convolutional networks,'' \emph{Journal of Systems Architecture}, vol.~95, pp. 9--18, 2019.

\bibitem{pang2017convolution}
Y.~Pang, M.~Sun, X.~Jiang, and X.~Li, ``Convolution in convolution for network in network,'' \emph{IEEE transactions on neural networks and learning systems}, vol.~29, no.~5, pp. 1587--1597, 2017.

\bibitem{agarap2018deep}
A.~F. Agarap, ``Deep learning using rectified linear units (relu),'' \emph{arXiv preprint arXiv:1803.08375}, 2018.

\bibitem{jacobs1991adaptive}
R.~A. Jacobs, M.~I. Jordan, S.~J. Nowlan, and G.~E. Hinton, ``Adaptive mixtures of local experts,'' \emph{Neural computation}, vol.~3, no.~1, pp. 79--87, 1991.

\bibitem{wang2020long}
X.~Wang, L.~Lian, Z.~Miao, Z.~Liu, and S.~X. Yu, ``Long-tailed recognition by routing diverse distribution-aware experts,'' in \emph{International Conference on Learning Representations}, 2021.

\bibitem{chen2023adamv}
T.~Chen, X.~Chen, X.~Du, A.~Rashwan, F.~Yang, H.~Chen, Z.~Wang, and Y.~Li, ``Adamv-moe: Adaptive multi-task vision mixture-of-experts,'' in \emph{Proceedings of the IEEE/CVF International Conference on Computer Vision}, 2023, pp. 17\,346--17\,357.

\bibitem{shazeer2017outrageously}
N.~Shazeer, A.~Mirhoseini, K.~Maziarz, A.~Davis, Q.~Le, G.~Hinton, and J.~Dean, ``Outrageously large neural networks: The sparsely-gated mixture-of-experts layer,'' in \emph{International Conference on Learning Representations}, 2017.

\bibitem{lin2017focal}
T.-Y. Lin, P.~Goyal, R.~Girshick, K.~He, and P.~Doll{\'a}r, ``Focal loss for dense object detection,'' in \emph{Proceedings of the IEEE international conference on computer vision}, 2017, pp. 2980--2988.

\bibitem{borse2023x}
S.~Borse, M.~Klingner, V.~R. Kumar, H.~Cai, A.~Almuzairee, S.~Yogamani, and F.~Porikli, ``X-align: Cross-modal cross-view alignment for bird's-eye-view segmentation,'' in \emph{Proceedings of the IEEE/CVF Winter Conference on Applications of Computer Vision}, 2023, pp. 3287--3297.

\bibitem{li2024bevformer}
Z.~Li, W.~Wang, H.~Li, E.~Xie, C.~Sima, T.~Lu, Q.~Yu, and J.~Dai, ``Bevformer: learning bird's-eye-view representation from lidar-camera via spatiotemporal transformers,'' \emph{IEEE Transactions on Pattern Analysis and Machine Intelligence}, 2024.

\bibitem{lang2019pointpillars}
A.~H. Lang, S.~Vora, H.~Caesar, L.~Zhou, J.~Yang, and O.~Beijbom, ``Pointpillars: Fast encoders for object detection from point clouds,'' in \emph{Proceedings of the IEEE/CVF conference on computer vision and pattern recognition}, 2019, pp. 12\,697--12\,705.

\bibitem{yin2021center}
T.~Yin, X.~Zhou, and P.~Krahenbuhl, ``Center-based 3d object detection and tracking,'' in \emph{Proceedings of the IEEE/CVF conference on computer vision and pattern recognition}, 2021, pp. 11\,784--11\,793.

\bibitem{chen2023focalformer3d}
Y.~Chen, Z.~Yu, Y.~Chen, S.~Lan, A.~Anandkumar, J.~Jia, and J.~M. Alvarez, ``Focalformer3d: focusing on hard instance for 3d object detection,'' in \emph{Proceedings of the IEEE/CVF International Conference on Computer Vision}, 2023, pp. 8394--8405.

\bibitem{zhang2024safdnet}
G.~Zhang, J.~Chen, G.~Gao, J.~Li, S.~Liu, and X.~Hu, ``Safdnet: A simple and effective network for fully sparse 3d object detection,'' in \emph{Proceedings of the IEEE/CVF Conference on Computer Vision and Pattern Recognition}, 2024, pp. 14\,477--14\,486.

\bibitem{vora2020pointpainting}
S.~Vora, A.~H. Lang, B.~Helou, and O.~Beijbom, ``Pointpainting: Sequential fusion for 3d object detection,'' in \emph{Proceedings of the IEEE/CVF conference on computer vision and pattern recognition}, 2020, pp. 4604--4612.

\bibitem{chen2023futr3d}
X.~Chen, T.~Zhang, Y.~Wang, Y.~Wang, and H.~Zhao, ``Futr3d: A unified sensor fusion framework for 3d detection,'' in \emph{proceedings of the IEEE/CVF conference on computer vision and pattern recognition}, 2023, pp. 172--181.

\bibitem{yin2021multimodal}
T.~Yin, X.~Zhou, and P.~Kr{\"a}henb{\"u}hl, ``Multimodal virtual point 3d detection,'' \emph{Advances in Neural Information Processing Systems}, vol.~34, pp. 16\,494--16\,507, 2021.

\bibitem{hao2024mbfusion}
X.~Hao, H.~Zhang, Y.~Yang, Y.~Zhou, S.~Jung, S.-I. Park, and B.~Yoo, ``Mbfusion: A new multi-modal bev feature fusion method for hd map construction,'' in \emph{2024 IEEE International Conference on Robotics and Automation (ICRA)}.\hskip 1em plus 0.5em minus 0.4em\relax IEEE, 2024, pp. 15\,922--15\,928.

\bibitem{liao2022maptr}
B.~Liao, S.~Chen, X.~Wang, T.~Cheng, Q.~Zhang, W.~Liu, and C.~Huang, ``Maptr: Structured modeling and learning for online vectorized hd map construction,'' in \emph{International Conference on Learning Representations}, 2023.

\bibitem{mustajbasic2025smab}
A.~Mustajbasic, S.~Chen, E.~Stenberg \emph{et~al.}, ``Smab: Simple multimodal attention for effective bev fusion,'' in \emph{2025 IEEE Intelligent Vehicles Symposium (IV)}.\hskip 1em plus 0.5em minus 0.4em\relax IEEE, 2025, pp. 1766--1772.

\bibitem{xia2025henet++}
Z.~Xia, Z.~Lin, Y.~Wang, and M.-H. Yang, ``Henet++: Hybrid encoding and multi-task learning for 3d perception and end-to-end autonomous driving,'' \emph{arXiv preprint arXiv:2511.07106}, 2025.

\bibitem{zhou2026unisparsebev}
H.~Zhou, Y.~Zhang, and H.~Qi, ``Unisparsebev: A multi-task learning framework with unified sparse query for autonomous driving,'' \emph{IEEE Transactions on Circuits and Systems for Video Technology}, 2026.

\bibitem{yang2025daocc}
Z.~Yang, Y.~Dong, J.~Wang, H.~Wang, L.~Ma, Z.~Cui, Q.~Liu, H.~Pei, K.~Zhang, and C.~Zhang, ``Daocc: 3d object detection assisted multi-sensor fusion for 3d occupancy prediction,'' \emph{IEEE Transactions on Circuits and Systems for Video Technology}, 2025.

\bibitem{loshchilov2017decoupled}
I.~Loshchilov and F.~Hutter, ``Decoupled weight decay regularization,'' in \emph{International Conference on Learning Representations}, 2019.

\bibitem{smith2019super}
L.~N. Smith and N.~Topin, ``Super-convergence: Very fast training of neural networks using large learning rates,'' in \emph{Artificial intelligence and machine learning for multi-domain operations applications}, vol. 11006.\hskip 1em plus 0.5em minus 0.4em\relax SPIE, 2019, pp. 369--386.

\end{thebibliography}

\end{document}